# Research Note
## Soft Goals Can Be Compiled Away

**Emil Keyder**                                                                 EMIL.KEYDER@UPF.EDU
*Universitat Pompeu Fabra*
*Roc Boronat, 138*
*08018 Barcelona Spain*

**Hector Geffner**                                                              HECTOR.GEFFNER@UPF.EDU
*ICREA & Universitat Pompeu Fabra*
*Roc Boronat, 138*
*08018 Barcelona Spain*

## Abstract

Soft goals extend the classical model of planning with a simple model of preferences. The best plans are then not the ones with least cost but the ones with maximum utility, where the utility of a plan is the sum of the utilities of the soft goals achieved minus the plan cost. Finding plans with high utility appears to involve two linked problems: choosing a subset of soft goals to achieve and finding a low-cost plan to achieve them. New search algorithms and heuristics have been developed for planning with soft goals, and a new track has been introduced in the International Planning Competition (IPC) to test their performance. In this note, we show however that these extensions are not needed: soft goals do not increase the expressive power of the basic model of planning with action costs, as they can easily be compiled away. We apply this compilation to the problems of the net-benefit track of the most recent IPC, and show that optimal and satisficing cost-based planners do better on the compiled problems than optimal and satisficing net-benefit planners on the original problems with explicit soft goals. Furthermore, we show that penalties, or negative preferences expressing conditions to avoid, can also be compiled away using a similar idea.

## 1. Models

A STRIPS problem is a tuple $P = \langle F, I, O, G \rangle$ where $F$ is a set of fluents, $I \subseteq F$ and $G \subseteq F$ are the initial state and goal situation, and $O$ is a set of actions or operators with precondition, add, and delete lists $Pre(a)$, $Add(a)$, and $Del(a)$ respectively, all of which are subsets of $F$. An action sequence $\pi = \langle a_0, \ldots, a_n \rangle$ is *applicable* in $P$ if the actions $a_i$, $i = 0, \ldots, n$, are all in $O$, and there exists a sequence of states $\langle s_0, \ldots, s_{n+1} \rangle$, such that $s_0 = I$, $Pre(a_i) \subseteq s_i$ and $s_{i+1} = s_i \cup Add(a_i) \setminus Del(a_i)$ for $i = 0, \ldots, n$. The applicable action sequence $\pi$ achieves a fluent $g$ if $g \in s_{n+1}$, and is a plan for $P$ if it achieves each goal $g$ in $G$, which we write as $\pi \models G$. In the classical setting, the cost of a plan $c(\pi)$ is given by $|\pi|$, the number of actions in $\pi$. This cost structure is generalized with the addition of a cost function over the operators:

KEYDER & GEFFNER**Definition 1** *A STRIPS problem with action costs is a tuple $P_c = \langle F, I, O, G, c \rangle$, where $P = \langle F, I, O, G \rangle$ is a STRIPS problem and $c$ is a function $c : O \mapsto \mathbb{R}_0^+$ where $\mathbb{R}_0^+$ stands for the non-negative reals.*

The cost of a plan $\pi$ for a problem $P_c$ is then given by

$$c(\pi) = \sum_{i=1}^{|\pi|} c(a_i) \qquad (1)$$

where $a_i$ denotes the $i$th action in $\pi$. The cost function $c(\pi) = |\pi|$ is obtained as a special case when $c(o) = 1$ for all $o \in O$. Adding *utilities* or *soft goals* to the problem formulation results in a new model:

**Definition 2** *A STRIPS problem with action costs and soft goals is a tuple $P_u = \langle F, I, O, G, c, u \rangle$, where $P = \langle F, I, O, G, c \rangle$ is a STRIPS problem with action costs, and $u$ is a partial function $u : F \mapsto \mathbb{R}^+$ that maps a subset of fluents (the soft goals) into positive reals.*

In a STRIPS problem with soft goals $P_u$, the *utility* of a plan is given by the difference between the total utility obtained by the plan and its cost:

$$u(\pi) = \sum_{p : \pi \models p} u(p) \ - \ c(\pi) \ . \qquad (2)$$

A plan $\pi$ for a problem with soft goals $P_u$ is optimal when no other plan $\pi'$ has a utility $u(\pi')$ higher than $u(\pi)$. The utility of an optimal plan for a problem with no hard goals is never negative, as the empty plan has non-negative utility and zero cost.

The most recent International Planning Competition (IPC6) featured *Sequential Optimal* and *Net Benefit Optimal* tracks in which the objective was to find optimal plans with respect to the models captured by Equation 1 and Equation 2 respectively (Helmert, Do, & Refanidis, 2008).[1]

## 2. Equivalence

Given a problem $P$ with soft goals, an equivalent problem $P'$ with action costs and *no soft goals* can be defined whose plans encode corresponding plans for $P$. This transformation, first introduced by Keyder and Geffner (2007), is simple and direct, yet seems to have escaped the attention of researchers in the area (Smith, 2004; Sanchez & Kambhampati, 2005; Bonet & Geffner, 2008; Baier, Bacchus, & McIlraith, 2007). Also, unlike the compilation of soft goals into numeric variables and arbitrary plan metrics (Edelkamp, 2006), the proposed transformation makes use of neither and requires from planners only the ability to handle

---

1. In PDDL3, soft goals are represented by expressions of the form (∗ $u$ (*is-violated* ⟨*pref*⟩)) appearing in the problem metric where *pref* is a preference or soft goal associated with a formula $A$. When $A$ is a single fluent, the expression corresponds to $u(A) = u$ in the terminology used here. Most of the competition benchmarks contain only preferences of this form. The more general case that arises when $A$ is a compound formula over fluents is considered in Section 4.

548

action costs, the basic functionality required in the satisficing track of the most recent IPC (Helmert et al., 2008).[2]

We write actions as tuples of the form $o = \langle Pre(o), \mathit{Eff}(o) \rangle$, where the effects can be positive (Adds) or negative (Deletes). We assume that for each soft goal fluent $p$, $P$ also contains a fluent $\bar{p}$ representing its negation. These can be introduced in the standard way, adding $\bar{p}$ to the initial state if $p$ is not initially true, and including $\bar{p}$ in the *Add* and *Delete* lists of all actions deleting or adding $p$ respectively (Gazen & Knoblock, 1997; Nebel, 2000). The problem $P'$ with action costs and no soft goals that is equivalent to the problem $P$ with soft goals can then be obtained by the following transformation:

**Definition 3** *For a STRIPS problem with action costs and soft goals $P = \langle F, I, O, G, c, u \rangle$, the compiled STRIPS problem with action costs is $P' = \langle F', I', O', G', c' \rangle$ with*

- $F' = F \cup S'(P) \cup \overline{S}(P) \cup \{\mathit{normal\text{-}mode}, \mathit{end\text{-}mode}\}$

- $I' = I \cup \overline{S}(P) \cup \{\mathit{normal\text{-}mode}\}$

- $G' = G \cup S'(P)$

- $O' = O'' \cup \{\mathit{collect}(p), \mathit{forgo}(p) \mid p \in SG(P)\} \cup \{\mathit{end}\}$

- $c'(o) = \begin{cases} c(o) & \text{if } o \in O'' \\ u(p) & \text{if } o = \mathit{forgo}(p) \\ 0 & \text{if } o = \mathit{collect}(p) \text{ or } o = \mathit{end} \end{cases}$

*where*

- $SG(P) = \{p \mid (p \in F) \wedge (u(p) > 0)\}$

- $S'(P) = \{p' \mid p \in SG(P)\}$

- $\overline{S}(P) = \{\overline{p'} \mid p' \in S'(P)\}$

- $\mathit{end} = \langle \{\mathit{normal\text{-}mode}\}, \{\mathit{end\text{-}mode}, \neg \mathit{normal\text{-}mode}\} \rangle$

- $\mathit{collect}(p) = \langle \{\mathit{end\text{-}mode}, p, \overline{p'}\}, \{p', \neg \overline{p'}\} \rangle$

- $\mathit{forgo}(p) = \langle \{\mathit{end\text{-}mode}, \overline{p}, \overline{p'}\}, \{p', \neg \overline{p'}\} \rangle$

- $O'' = \{\langle Pre(o) \cup \{\mathit{normal\text{-}mode}\}, \mathit{Eff}(o) \rangle \mid o \in O\}$

---

2. Edelkamp's transformation associates with the soft goals $p_1, \ldots, p_m$ numeric variables $n_1, \ldots, n_m$, each with domain $\{0, 1\}$. The utility for a plan is then expressed as $U(\pi) = \sum_{i=1}^{n} n_i * u(p_i) - cost(\pi)$, where $u(p_i)$ represents the utility associated with soft goal $p_i$ and $n_i$ represents the value of the numeric variable in the final state achieved by the plan. This transformation also eliminates soft goals, but requires in its place a plan metric whose terms (namely, whether the variables $u(p_i)$ are 1 or 0) are state-dependent. Current heuristics can not deal with such metrics (See Sections 3 and 4).

For each soft goal $p$ in $P$, the transformation adds a dummy hard goal $p'$ in $P'$ that can be achieved in two ways: with the action *collect(p)* that has cost 0 but requires $p$ to be true, or with the action *forgo(p)* that has cost equal to the utility of $p$ yet can be performed when $p$ is false, or equivalently when $\bar{p}$ is true. These two actions can be used only after the *end* action that makes the fluent *end-mode* true, while the actions from the original problem $P$ can be used only when the fluent *normal-mode* is true prior to the execution of the *end* action. Moreover, exactly one of $\{collect(p), forgo(p)\}$ can appear for each soft goal $p$ in the plan, as both delete their shared precondition $\overline{p'}$, which no action makes true. As there is no way to make *normal-mode* true again after it is deleted by the *end* action, all plans $\pi'$ for $P'$ have the form $\pi' = \langle \pi, end, \pi'' \rangle$, where $\pi$ is a plan for $P$ and $\pi''$ is a sequence of $|S'(P)|$ *collect(p)* and *forgo(p)* actions in any order, the former appearing when $\pi \models p$, and the latter otherwise.

The two problems $P$ and $P'$ are equivalent in the sense that there is a correspondence between the plans for $P$ and $P'$, and corresponding plans are ranked in the same way. More specifically, for any plan $\pi$ for $P$, there is a plan $\pi'$ for $P'$ that extends $\pi$ with the *end* action and a set of *collect* and *forgo* actions, and this plan has cost $c(\pi') = -u(\pi) + \alpha$, where $\alpha$ is a constant that is independent of both $\pi$ and $\pi'$. Finding an optimal (maximum utility) plan $\pi$ for $P$ is therefore equivalent to finding an optimal (minimum cost) plan $\pi'$ for $P'$.

**Proposition 1 (Correspondence between plans)** *For an applicable action sequence $\pi$ in $P$, let an* extension *$\pi'$ of $\pi$ denote any sequence obtained by appending to $\pi$ the end action followed by some permutation of the actions collect(p) and forgo(p) for all $p \in SG(P)$, when $\pi \models p$ and $\pi \not\models p$ respectively. Then*

$$\pi \text{ is a plan for } P \iff \pi' \text{ is a plan for } P'$$

*Proof:* ($\Rightarrow$) The new actions in $P'$ do not delete any $p \in F$, so any hard goal achieved by $\pi$ will remain true in the final state reached by $\pi'$, and we have that $\pi' \models G$. For all $p \in F$ such that $u(p) > 0$, either $\pi \models p$ or $\pi \not\models p$. In the first case, $p'$ is achieved by *collect(p)*, in the second, by *forgo(p)*, therefore $\pi' \models S'(P)$. Since $G' = G \cup S'(P)$, we have that $\pi' \models G'$.
($\Leftarrow$) If $\pi'$ is a plan for $P'$, then all hard goals $G$ in $P$ must be made true by $\pi'$ before the *end* action, as after this action only *collect* and *forgo* actions can be applied and these can not make any $p \in F$ true. The plan obtained by removing the *end* action and all *collect* and *forgo* actions must therefore achieve $G$ and thus is a valid plan for $P$. □

**Proposition 2 (Correspondence between utilities and costs)** *Let $\pi_1$ and $\pi_2$ be two plans for $P$, and let $\pi'_1$ and $\pi'_2$ be extensions of $\pi_1$ and $\pi_2$ respectively. Then,*

$$u(\pi_1) > u(\pi_2) \iff c(\pi'_1) < c(\pi'_2)$$

*Proof:* Let $\pi$ be a plan for $P$ and $\pi'$ an extension of $\pi$. We demonstrate that $c(\pi') = -u(\pi) + \sum_{p \in SG(P)} u(p)$. Since the summation in this expression is a constant for a given problem $P$, the assertion follows directly:

$$\begin{aligned}
c(\pi') &= c(\pi) + c'(end) + \sum_{forgo(p) \in \pi'} c'(forgo(p)) + \sum_{collect(p) \in \pi'} c'(collect(p)) \\
&= c(\pi) + \sum_{forgo(p) \in \pi'} c'(forgo(p)) \\
&= c(\pi) + \sum_{p:\pi \not\models p} u(p) \\
&= c(\pi) + \sum_{p \in SG(P)} u(p) - \sum_{p:\pi \models p} u(p) \\
&= -u(\pi) + \sum_{p \in SG(P)} u(p)
\end{aligned}$$

$\square$

**Proposition 3 (Equivalence)** *Let $\pi$ be a plan for $P$, and $\pi'$ be a plan for $P'$ that extends $\pi$. Then,*

$$\pi \text{ is an optimal plan for } P \iff \pi' \text{ is an optimal plan for } P'$$

*Proof:* Direct from the two propositions above. $\square$

In the following section, we empirically compare the performance of net-benefit planners on problems $P$ with explicit soft goals to that of sequential planners on problems $P'$ in which soft goals have been compiled away. In order to improve the latter, we make the transformation of Definition 3 more effective with a simple trick. Recall that for a single plan $\pi$ for $P$, there are many extensions $\pi'$ in $P'$, all containing the same actions and having the same cost, but differing in the way the *collect* and *forgo* actions are ordered. For efficiency purposes, the implementation enforces a fixed but arbitrary ordering $p_1, \ldots, p_m$ on the soft goals in $P$ by adding the dummy hard goal $p'_i$ as a precondition of the actions $collect(p_{i+1})$ and $forgo(p_{i+1})$ for $i = 1, \ldots, m-1$. The result is that there is a single possible extension $\pi'$ of every plan $\pi$ in $P$, and the space of plans to search is therefore reduced. This optimization is used in the experiments reported below.

## 3. Experimental Results

The formal results above imply that the best plans for a problem $P$ with action costs and soft goals can be computed by looking for the best plans for the compiled problem $P'$ with action costs and no soft goals, to which standard classical planning techniques can be applied. To test the practical value of the transformation, we evaluate the performance of both optimal and satisficing planning techniques for soft goals. Some problems in the test suite contain preferences over conjunctions rather than single fluents. Such preferences are handled with a variant of the approach described above, detailed in Section 4.

The results shown in the three columns in Table 1 labelled 'Net-benefit optimal planners' are the results as reported by the organizers of the 2008 International Planning Competition (IPC6) (Helmert et al., 2008). All other results were obtained using the same machines and settings as used in the competition: Xeon Woodcrest computers with clock speeds of 2.33 GHz, with a time limit of 30 minutes and a memory limit of 2GB.

|  | Net-benefit optimal planners | | | Sequential optimal planners | | | |
| --- | --- | --- | --- | --- | --- | --- | --- |
| Domain | Gamer | $\text{HSP}_\text{P}^*$ | Mips-XXL | Gamer | $\text{HSP}_\text{F}^*$ | $\text{HSP}_0^*$ | Mips-XXL |
| crewplanning(30) | 4 | 16 | 8 | - | 8 | **21** | 8 |
| elevators (30) | 11 | 5 | 4 | **19** | 8 | 8 | 3 |
| openstacks (30) | **7** | 5 | 2 | 6 | 4 | 6 | 1 |
| pegsol (30) | 24 | 0 | 23 | 22 | **26** | 14 | 22 |
| transport (30) | 12 | 12 | 9 | - | **15** | **15** | 9 |
| woodworking (30) | 13 | 11 | 9 | - | 10 | **14** | 7 |
| total | 71 | 49 | 55 | | 71 | **78** | 50 |

Table 1: Coverage for optimal planners: The leftmost three columns give the number of problems solved by each of the planners in the Net Benefit Optimal track of IPC6, as reported by the competition organizers. The rightmost four columns give the number of *compiled* problems solved by the Sequential Optimal versions of these planners. Dashes indicate that the version of the planner could not be run on that domain.

In the first set of experiments, we consider the problems used in the *Net Benefit Optimal* (NBO) track of IPC6, in which soft goals are defined in terms of *goal-state preferences* (Gerevini & Long, 2006), and compare the results obtained by the three optimal net-benefit planners with the results obtained by their *Sequential Optimal* (SO) variants on their compilations.[3] The three planners entered in the NBO track of IPC6 were Gamer, Mips-XXL, and $\text{HSP}_\text{P}^*$. The SO planners we test on the compiled versions of the NBO problems are the SO versions of Gamer (Edelkamp & Kissmann, 2008) and Mips-XXL (Edelkamp & Jabbar, 2008) and the two SO planners $\text{HSP}_\text{F}^*$ and $\text{HSP}_0^*$ (Haslum, 2008).[4] These were ranked first, fifth, second, and third, respectively, in the SO track (Helmert et al., 2008). Three out of the six domains from the NBO track of IPC6 involve numeric variables that appear in the preconditions of actions. The SO version of Gamer does not handle numeric variables, and we are therefore unable to run Gamer on such problems. Numeric variables never appear as soft goals and are left untouched by our compilation.

The data in Table 1 show that the two $\text{HSP}^*$ planners from the SO track run on the compiled problems do as well as, or better than, the best planner from the NBO track run on the original problems with soft goals. The maximum number of solved problems for a domain is higher for the NBO track planners in only a single domain, openstacks (7 vs. 6). In all other domains, SO planners are able to solve a larger number of problems than the maximum number solved by any NBO planner. Considering the performance of the NBO and SO variants of each planner, the compilation benefits most the two versions of the heuristic search planner HSP*, leaving the BDD planners Gamer and Mips-XXL relatively unaffected. Interestingly, HSP$_0^*$ using the compilation ends up solving more problems than Gamer, the winner of the NBO track (78 vs. 71). The drastically better performance of the SO versions of HSP* compared to the net-benefit version is the result of the simple scheme for handling soft goals in the latter, in which optimal plans are computed for each possible subset of soft goals in the problem (roughly), and a change in the search algorithm from IDA* to A*.

---

3. The compiled problems are currently available at http://ipc.informatik.uni-freiburg.de/Domains.
4. All versions of $\text{HSP}^*$ have a bug which may cause suboptimal or invalid solutions to be computed in domains with non-monotonic numeric variables (numeric variables whose values may both increase and decrease) that occur in preconditions of actions or goals (See http://ipc.informatik.uni-freiburg.de/Planners). Such variables are present only in the *transport* domain out of all those tested, yet plans computed by $\text{HSP}^*$ for both versions of the domain turn out to be valid (as verified by the VAL plan validator, Howey & Long, 2003) and optimal in the instances in which they can be checked against the costs of plans computed by other planners.

|  | Net-benefit satisficing planners | | | Cost satisficing planners |
|---|---|---|---|---|
| Domain | SGPlan | YochanPS | Mips-XXL | Lama |
| elevators (30) | 0 | 0 | 8 | **23** |
| openstacks (30) | 2 | 0 | 2 | **28** |
| pegsol (30) | 0 | 5 | 23 | **29** |
| rovers (20) | 8 | 2 | 1 | **17** |
| total | 10 | 7 | 34 | **97** |

Table 2: Coverage and quality for satisficing planners: The entries indicate the number of problems for which the planner generated the best quality plan.

In the second set of experiments, we consider the three domains from the NBO track of IPC6 which do not contain numeric variables in the preconditions of actions, and the domain *rovers* from the net-benefit track of IPC5. Domains containing numeric variables in the preconditions of actions are not considered due to the lack of state-of-the-art cost-based planners able to handle them. Domains other than *rovers* from the NB track of IPC5 are not considered as they contain disjunctive, existentially qualified, or universally qualified soft goals which our current implementation does not support. The satisficing net-benefit planners we test on these problems are SGPlan (Hsu & Wah, 2008), the winner of the net benefit track from IPC5, YochanPS (Benton, Do, & Kambhampati, 2009), which received a distinguished performance award in the same competition, and a satisficing variant of MIPS-XXL, which also received a distinguished performance award in that competition and competed in the optimal track of IPC6. We solve the compiled versions of the problems with LAMA, the winner of the sequential satisficing track from IPC6. YochanPS, MIPS-XXL, and LAMA are anytime planners, and the results discussed below refer to the cost of the best plan found by each at the end of the evaluation period of 30 minutes.

Entries in Table 2 show the number of problems in each domain for which the plan generated by a planner is the best or only plan produced. We report this data rather than showing graphs of plan utilities as the absolute difference between the quality of plans is not meaningful in itself except when the shortest plans (that ignore costs and/or soft goals) for the problem are significantly more costly. The results show that running a state-of-the-art cost-based planner on the compiled problems yields the best plan in 98 out of the total 110 instances, almost three times the number of instances in which the best-performing native soft goals planner, MIPS-XXL, gives the best plan. Furthermore, in 22 out of the 23 problems for which MIPS-XXL finds the best plan in the *pegsol* domain, LAMA finds a plan with the same quality. The problems in which satisficing net-benefit planners outperform LAMA run on the compiled problems are therefore very few.

These results appear to contradict the results reported by Benton et al. (2009), where the native net-benefit planner, Yochan[PS], yields better results than a cost-based planner, Yochan[COST], run on problems compiled according to an earlier version of our transformation (Keyder & Geffner, 2007). The discrepancy appears to be the result of the non-informative cost-based heuristic used in Yochan[COST], which leads to plans that forgo all soft goals, and the fact that they do not make use of the optimization discussed at the end of Section 2, which results in an unnecessary blowup of the state space. For an analysis of the differences between some recent cost-based planners, see the paper by Keyder and Geffner (2008).

## 4. Extensions

We have shown that it is possible to compile away positive utilities $u(p)$ associated with single fluents $p$. We show now that this compilation can be extended to deal with positive utilities defined on *formulas over fluents* and to *negative utilities* defined on both single fluents and formulas. Negative utilities stand for conditions to be avoided rather than sought; for example, a utility $u(p \wedge q) = -10$ penalizes a plan that results in a state where both $p$ and $q$ are true with an extra cost of 10. The compilation of soft goals defined on formulas is based on the standard compilation of goal and precondition formulas in classical planning (Gazen & Knoblock, 1997; Nebel, 1999).

A positive utility on a logical formula $A$ can be compiled away by introducing a new fluent $p_A$ that can be achieved at zero cost from any end state where $A$ holds, and by assigning the utility associated with $A$ to $p_A$. If $A$ is a DNF formula $D_1 \vee \ldots \vee D_n$, it suffices to add $n$ new actions $a_1, \ldots, a_n$ with $a_i = \langle D_i, p_A \rangle$ for $i = 1, \ldots, n$. If $A$ is a CNF formula $C_1 \wedge \ldots \wedge C_n$, a fluent $p_i$ is introduced for each $i = 1, \ldots, n$, along with actions $a_{ij} = \langle C_{ij}, p_i \rangle$ for $j = 1, \ldots, |C_i|$, where $C_{ij}$ stands for the $j$th fluent of $C_i$. We also introduce an action $a = \langle \{p_1, \ldots, p_n\}, p_A \rangle$ that allows the addition of fluent $p_A$ in states where $A$ holds. All the newly introduced actions have zero cost, and must be applicable in $P'$ after the actions of the original problem $P$ and before the *collect* and *forgo* actions. The best extensions of any plan $\pi$ that achieves $A$ in $P$ will then achieve $p_A$ and use the *collect* action to achieve the hard goal fluent $p'_A$ associated with $p_A$ at zero cost.

A negative utility $u(A) < 0$ on a formula $A$ in DNF or CNF can be compiled away in two steps, by first substituting a positive utility $-u(\neg A)$ on the negation $\neg A$ of $A$ and then compiling this positive utility on a formula into a utility on a single fluent as described above. This makes use of the fact that the negation of a formula in CNF is a formula in DNF and vice versa.

## 5. Summary

We have shown that soft goals do not add expressive power and can be easily compiled away. This implies that no new search algorithms or heuristics are strictly required for handling them. From a practical standpoint, experiments indicate that state-of-the-art sequential planners outperform state-of-the-art net-benefit planners on compiled versions of the benchmarks used in recent planning competitions. Furthermore, similar transformations can be used to compile away positive and negative utilities on logical formulas in DNF or CNF.

## Acknowledgments

We thank Malte Helmert for his help with compiling and running many of the IPC6 planners, Patrik Haslum for his help with all aspects of various versions of HSP, and J. Benton for his help with compiling and running Yochan$^{PS}$. H. Geffner is partially supported by Grant TIN2006-15387-C03-03 from MEC, Spain.